%% file: main.tex
\documentclass[10pt,twocolumn,letterpaper]{article}
\usepackage{graphicx}
\usepackage{multirow}
\usepackage[cvprfinal]{cvpr}      

\definecolor{brandeisblue}{rgb}{0.0, 0.44, 1.0}

\newcommand{\mypara}[1]{\noindent \textbf{#1}\hspace{0.1in}}

\newcommand{\myname}[1]{LiteVLM{}}
\definecolor{cvprblue}{rgb}{0.21,0.49,0.74}
\usepackage[pagebackref,breaklinks,colorlinks,allcolors=cvprblue]{hyperref}

\def\paperID{*****} 
\def\confName{CVPR}
\def\confYear{2025}

\title{LiteVLM: A Low-Latency Vision-Language Model Inference Pipeline for Resource-Constrained Environments}

\author{Jin Huang, Yuchao Jin, Le An, Josh Park\\
NVIDIA\\
{\tt\small \{jinhu, yuchaoj, lean, joshp\}@nvidia.com}
}

\begin{document}
\maketitle
\begin{abstract}
This paper introduces an efficient Vision-Language Model (VLM) pipeline specifically optimized for deployment on embedded devices, such as those used in robotics and autonomous driving. The pipeline significantly reduces the computational overhead by jointly leveraging patch selection to filter irrelevant camera views, a token selection module to reduce input sequence length for the LLM, and speculative decoding to accelerate token generation. Evaluation on the NVIDIA DRIVE Thor platform for automonous driving application, our pipeline achieves $2.5\times$ end-to-end latency reduction without compromising task accuracy. The speed-up further increases to $3.2\times$ when applying FP8 post-training quantization. These results demonstrate our pipeline as a viable solution for enabling real-time VLM deployment in resource-constrained environments.
\end{abstract}
\input{intro}    
\input{related}
\input{design}

\input{impl}
\input{evaluation}
{
    \small
    \bibliographystyle{ieeenat_fullname}
    \bibliography{main}
}


\end{document}

%% file: intro.tex
\section{Introduction}
Multi-modal Large Language Models (MLLMs), particularly Vision-Language Models (VLMs), have demonstrated remarkable capabilities in visual understanding and reasoning~\cite{Shi2024EagleET, llava, qwenvl, minigpt4, internvl, gpt4v, Qwen2VL}, .
Despite their strength, VLMs are computationally intensive. A typical architecture includes a Vision Transformer (ViT)~\cite{ViT} as encoder, an alignment module (e.g., MLP or Q-Former~\cite{BLIP2} structure) to align visual and text tokens, and an LLM decoder. In addition, many state-of-the-art VLMs also preprocess images of dynamic resolutions into fine-grained fixed-size patches~\cite{Qwen2VL} to boost performance but with increased computation. 
In turn, challenges remain for real-time deployment on resource-constrained embedded hardware such as in robotics and autonomous driving, where inference latency becomes a critical bottleneck, as rapid decision making is essential in those applications. 


To address VLM latency, visual token compression~\cite{fastv, Prumerge, TokenPacker, AVGLLaVA, M3, Tokenflex} reduces input sequence length by merging, pooling, or pruning less informative tokens, which can mitigate the quadratic computational cost of self-attention layers in LLMs. Since visual tokens often dominate the input sequence, such compression can lead to significant computational savings.
Another widely used technique is speculative decoding~\cite{speculative, medusa, eagle, Eagle2}, accelerating the decoding stage by using a smaller draft model to propose multiple candidate tokens, which are then verified by the primary LLMs. This method significantly reduces the generation latency by potentially accepting multi-tokens per iteration compared with auto-regressive manner. 

However, these methods by their own may not be sufficient to meet latency requirement for production environment. This paper introduces a VLM pipeline specifically engineered for efficient deployment on embedded systems. Our key contributions are: 
\begin{enumerate}
    \item We propose an efficient VLM architecture \textbf{LiteVLM} by jointly leveraging patch selection, token selection, and speculative decoding.  
    \item We benchmark our method for autonomous driving application on NVIDIA DRIVE Thor platform and demonstrate $\mathbf{2.5\times}$ reduction in latency compared with baseline while maintaining accuracy and even further speed up with FP8 quantization.
\end{enumerate}

%% file: related.tex
\section{Related Work}
\begin{figure*}[htbp]
    \vspace{-0.2cm}
    \centering
    \includegraphics[width=0.85\linewidth]{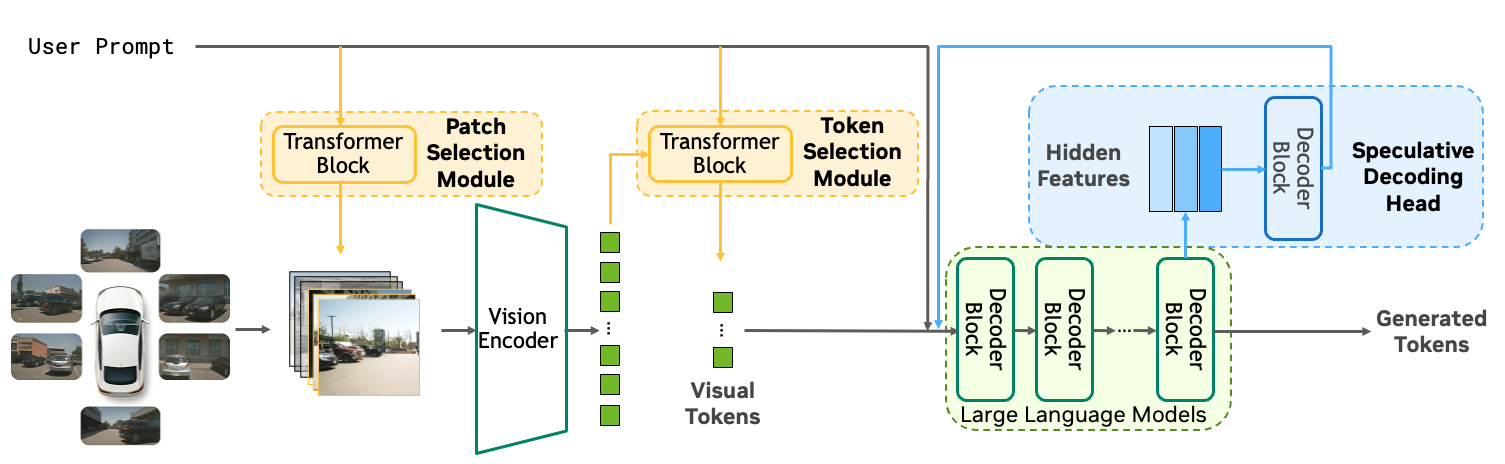}
    \caption{The proposed framework builds upon a Vision-Language Model by introducing two novel modules: the \textbf{Patch Selection Module} and the \textbf{Token Selection Module} (highlighted in \textcolor{orange}{Orange}). We also incorporate \textbf{Speculative Decoding Head} (highlighted in \textcolor{blue}{Blue}) to accelerate the decoding process. Together, these components enable efficient token generation for real-time applications.}
    \label{fig:overview}
    \vspace{-0.5cm}
\end{figure*}
\subsection{Visual Token Compression}
In VLMs, visual tokens frequently constitute the majority of the input sequence due to the fine-grained encoding of input images. This results in substantial computational overhead, particularly during the prefill stage, where the model processes the entire sequence before producing any output. To mitigate this issue, a range of visual token compression techniques have been proposed~\cite{fastv, Prumerge, TokenPacker, AVGLLaVA, M3, Tokenflex}. Some methods are based on \textbf{Token Merging} and reduce the token count by combining adjacent tokens or pooling their features~\cite{qwenvl, internvl}. More advanced methods use adaptive merging based on routing to achieve variable compression rates~\cite{M3, AVGLLaVA, Tokenflex}. However, these techniques often require finetuning the VLM to adapt to the merged features. Another strategy is \textbf{Token Pruning} which aims to identify and remove less informative tokens based on importance scores~\cite{Prumerge, ToSA}. For example, FastV~\cite{fastv} and HiRED~\cite{HiRED} prunes tokens based on self-attention scores from LLM or ViT encoder, which can avoid extensive model fine-tuning. While achieving good performance, these methods pose a risk of potentially discarding tokens associated with critical objects (e.g., distant pedestrians, traffic signs) that may receive low attention scores yet remain semantically crucial.

\subsection{Vision-Language Models on Edge Devices}
Recent progress in multi-modal learning has led to the development of compact, high-performing open-source Vision-Language Models (VLMs), making them increasingly viable for deployment on resource-constrained edge devices.
Notable examples include QwenVL-1.5B/3B~\cite{qwenvl, Qwen2VL} and InternVL-1B/2B~\cite{internvl}. These compact VLMs are increasingly being adopted by researchers in robotics and autonomous driving, domains where on-device deployment is often essential. For instance, studies such as~\cite{Dolphins, DriveGPT4, drivevlm, omnidrive} have finetuned these models on driving-specific datasets for tasks including scene understanding, perception, behavior prediction, and trajectory generation.

However, despite their reduced size compared to larger counterparts, these models, while easily manageable on high-end server GPUs, often still exceed the computational and memory limits of typical embedded devices. Consequently, real-world deployment remains a significant challenge, primarily due to stringent latency requirements and limited on-device resources.

%% file: design.tex
\section{Design}
\input{table}
Figure~\ref{fig:overview} shows the overview of \textbf{\myname{}} pipeline for an autonomous driving use case with multi-view inputs. Building upon a standard VLM, we introduce three key enhancements: 1. Patch Selection Module to reduce ViT input and subsequent LLM prefill latency by filtering irrelevant camera views, 2. Token Compression Module to further decrease LLM prefill cost, and 3. Speculative Decoding Head to accelerate autoregressive token generation.

\mypara{Patch Selection Module} Current VLMs often process visual input by dividing it into a fixed grid of patches~\cite{mini_internvl}. This query-agnostic approach can be inefficient, as many queries pertain only to specific camera views, resulting in unnecessary computation for both vision encoder and LLM. 



To address this, the proposed Patch Selection Module dynamically identifies relevant camera views based \textit{only} on the input text query before image encoding. The core idea is that query semantics often imply spatial focus. The module consists of a small transformer encoder with $4$ layers. The input text query is first tokenized and then fed into the transformer encoder. To determine view relevance, we employ a cross-attention mechanism: the encoded query features attend to $N$ learnable latent query vectors, where $N$ is the number of cameras in a multi-view perception setting. The resulting attention scores for each view-specific latent query are then mapped to independent logits, effectively treating the selection as $N$ binary classifications.


This module is trained independently prior to VLM fine-tuning, using the sum of Binary Cross-Entropy (BCE) losses across the 
$N$ views as the training objective.
Ground-truth labels are generated using a hybrid strategy.
Generating the ground-truth labels involves a hybrid approach. 
For queries that include explicit view references (e.g., “front left”), we apply heuristic lexical matching. For more implicit queries, we leverage GPT-3.5-Turbo as an evaluator: the model is prompted with the query and tasked with identifying the camera views required to answer it.

At inference time, the final scores will be a weighted sum from lexical mapping and the logits from this module. The patches from different views with scores above a predefined threshold are selected for the subsequent ViT encoder.

\mypara{Token Selection Module:} Following patch selection, the sequence of visual tokens fed into the LLM can still be substantial. Inspired by methods like FastV~\cite{fastv}, we introduce Token Selection Module by extracting the first decoder layer of our finetuned LLM and instantiating it as a lightweight, standalone Token Selection Module. This architectural separation allows us to finetune this module specifically for the task of token pruning, independent of the main VLM's weights after its initial finetuning.



A key challenge is generating appropriate training signals for this module. We create synthetic ground truth labels using a two-pronged strategy: 1) leveraging VLM self-attention hidden features as importance scores of visual tokens, then 2)  preserving critical objects by assigning high importance to their visual tokens that correspond to critical objects such as pedestrians and vehicles, identified using bounding boxes from the nuScenes dataset~\cite{nuscenes}. This standalone design offers a significant deployment advantage by preserving the integrity of the LLM's execution graph and simplifying integration with optimized inference engines.

\mypara{Speculative Decoding Head}
To accelerate the token generation phase, we integrate speculative decoding based on the Eagle-2 methodology~\cite{Eagle2}. We instantiate a lightweight draft model using a single decoder layer. During generation, this draft model takes the hidden state features from the final decoder layer of the VLM as input. It then rapidly generates a short sequence of candidate tokens. These candidate tokens are subsequently verified in one forward pass with KV cache, by the language models in VLM, allowing multiple tokens to be accepted per decoding step, thereby reducing overall generation latency.

Our design can also benefit other VQA or VLM tasks by accelerating VLM deployment. For generic VQA tasks, the spatial information may not be present. In this case, the patch selection module can adopt alternative design such as relying more on LLM hidden features.


\begin{figure}[ht]
    \vspace{-0.4cm}
    \centering
    \includegraphics[width=0.8\linewidth]{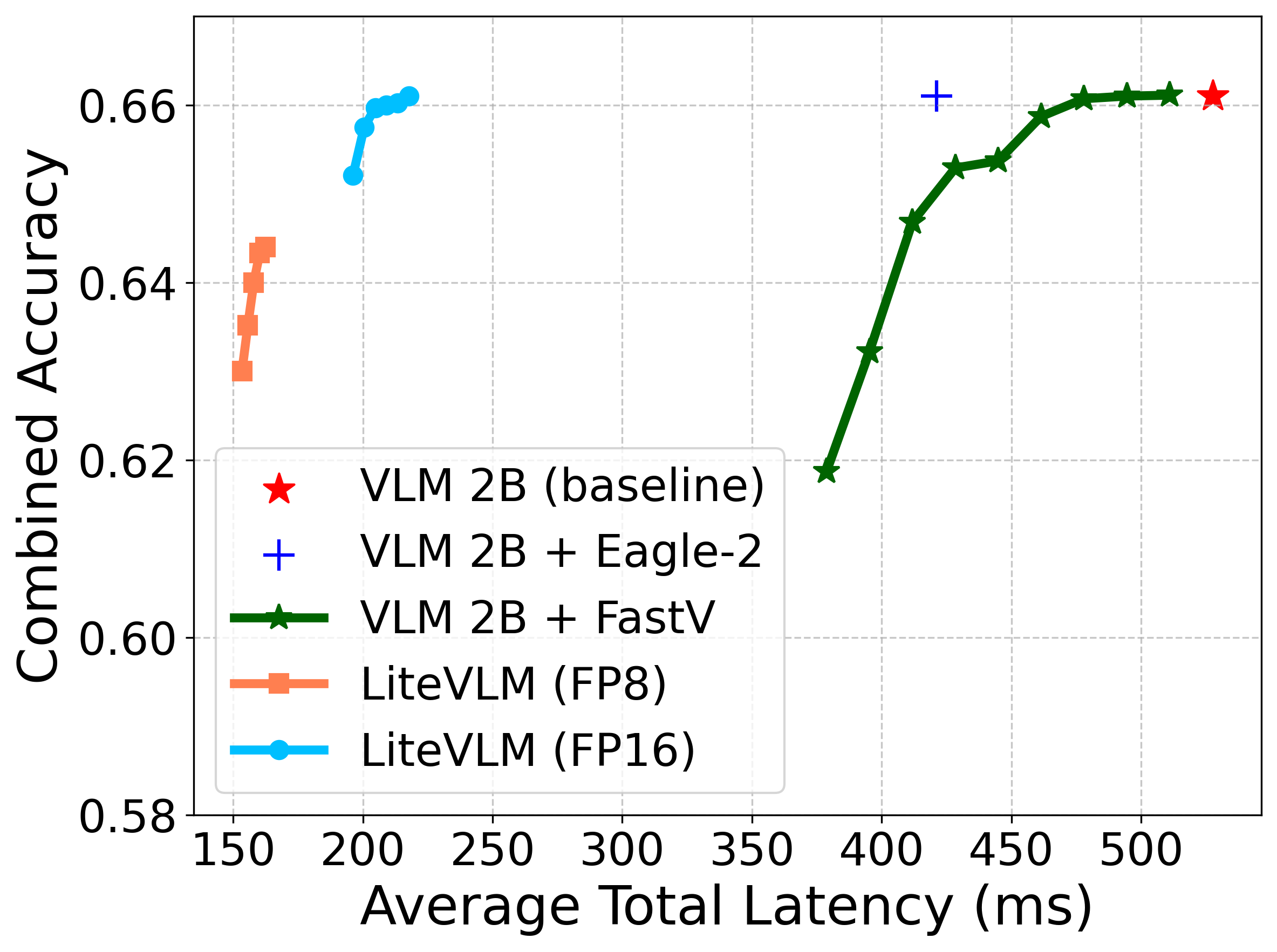}
    \vspace{-0.3cm}
    \caption{Average end-to-end latency of our proposed pipeline compared to baseline 2B VLMs across different configurations.}
    \label{fig:avg-total-latency}
    \vspace{-0.5cm}
\end{figure}

\begin{figure*}[htbp]
    \vspace{-0.2cm}
    \centering
     \begin{subfigure}[t]{0.30\textwidth}
         \centering
         \includegraphics[width=\textwidth, keepaspectratio]{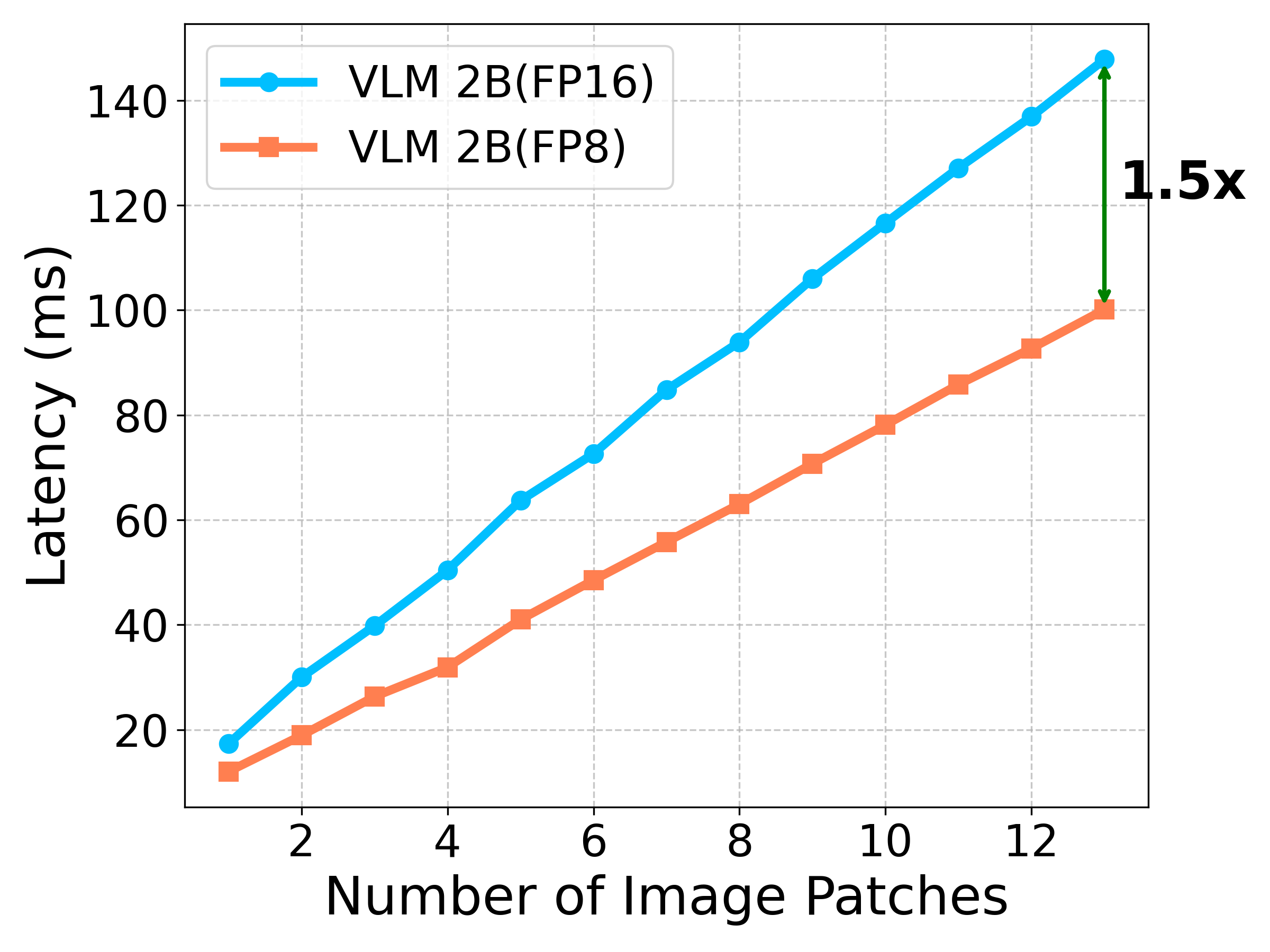}
         \vspace{-0.5cm}
         \caption{Vision Encoder Latency}
         \label{fig:vision-encoder-latency}
     \end{subfigure}
     \begin{subfigure}[t]{0.30\textwidth}
         \centering
         \includegraphics[width=\textwidth, keepaspectratio]{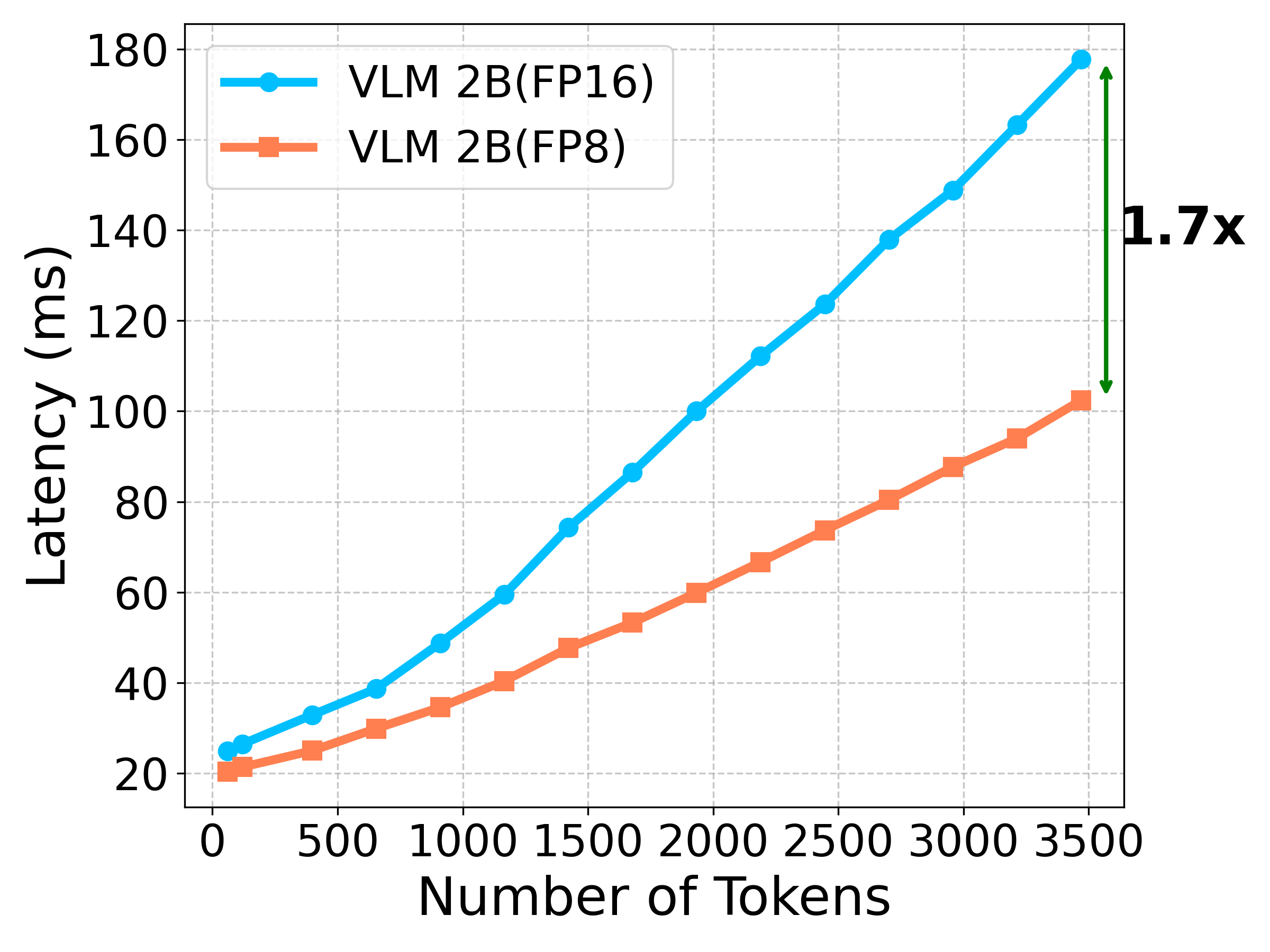}
         \vspace{-0.5cm}
         \caption{LLM Prefill stage Latency}
         \label{fig:first-token-latency}
     \end{subfigure}
     \begin{subfigure}[t]{0.30\textwidth}
         \centering
         \includegraphics[width=\textwidth, keepaspectratio]{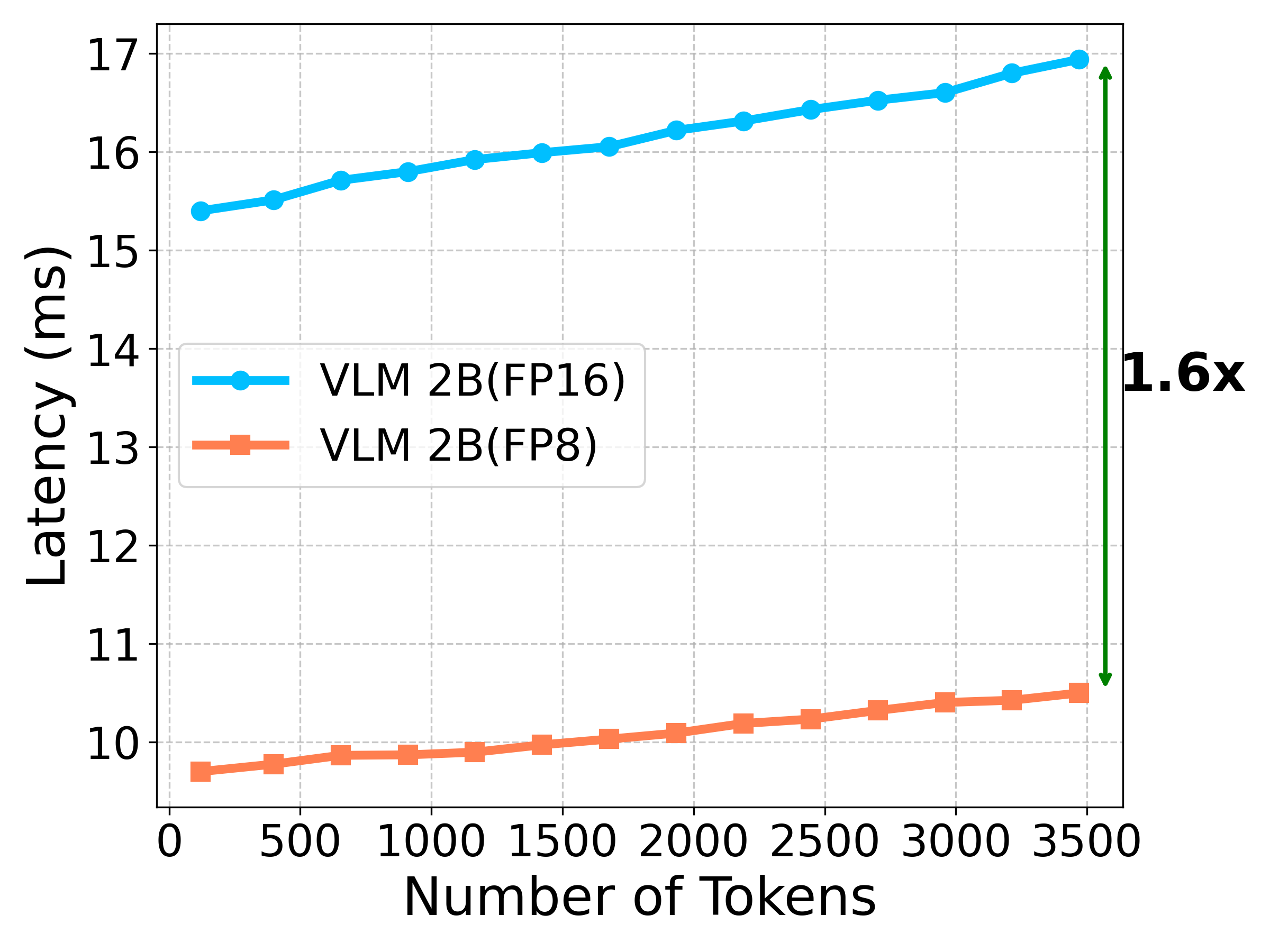}
         \vspace{-0.5cm}
         \caption{LLM Extend-One-Token Latency}
         \label{fig:extend-one-latency}
     \end{subfigure}
    \vspace{-0.3cm}
    \caption{The latency of different stages of VLM 2B of FP16 and FP8 execution on NVIDIA DRIVE Thor Platform}
        \label{fig:latency-breakdown}
    \vspace{-0.5cm}
\end{figure*}

%% file: table.tex
\begin{table*}[htbp]
\centering
\footnotesize
\caption{Inference latency and accuracy of different execution stages on NVIDIA DRIVE Thor platforms. All models are deploy in FP16 by default. We also benchmarked our model in FP8 precision.}
\vspace{-0.2cm}
\label{table:overall-performance}
\begin{tabular}{@{}l|cc|cc|cc|c|ccc@{}}
\toprule
\multirow{2}{*}{\textbf{Model}} &
  \multicolumn{2}{c|}{\textbf{Vision}} &
  \multicolumn{2}{c|}{\textbf{LLM Prefill}} &
  \multicolumn{2}{c|}{\textbf{LLM Decode}} &
  \multirow{2}{*}{\begin{tabular}[c]{@{}c@{}}Selection\\ Module\\ Latency\\ (ms)\end{tabular}} &
  \multicolumn{3}{c}{\textbf{End-to-End}} \\
 &
  \begin{tabular}[c]{@{}c@{}}Avg. \\ \#Image\\ Patches\end{tabular} &
  \begin{tabular}[c]{@{}c@{}}ViT \\ Latency \\ (ms)\end{tabular} &
  \begin{tabular}[c]{@{}c@{}}Avg. \\ \#Input\\ Tokens\end{tabular} &
  \begin{tabular}[c]{@{}c@{}}Prefill \\ Latency \\ (ms)\end{tabular} &
  \begin{tabular}[c]{@{}c@{}}Avg.\\ Extend-One\\ Latency (ms)\end{tabular} &
  \begin{tabular}[c]{@{}c@{}}Decoding\\ Latency \\ (ms)\end{tabular} &
   &
  \begin{tabular}[c]{@{}c@{}}Total \\ Latency \\ (ms)\end{tabular} &
  \multicolumn{1}{l}{Speed-Up} &
  \begin{tabular}[c]{@{}c@{}}Combined\\ Accuracy\end{tabular} \\ \midrule
VLM 2B              & 12  & 136.9 & 3214 & 163.2 & 16.9 & 229.6 & -    & 529.7 & $1.0\times$ & 0.6618 \\ \midrule
VLM 2B + FastV~\cite{fastv}(r=0.7) & 12  & 136.9 & 2292 & 117.9 & 16.4 & 222.8 & -    & 477.6 & $1.1\times$ & 0.6610 \\ \midrule
VLM 2B + FastV~\cite{fastv}(r=0.3) & 12  & 136.9 & 1063 & 59.5  & 15.9 & 216.1 & -    & 412.5 & $1.3\times$ & 0.6468 \\ \midrule
VLM 2B + Eagle~\cite{eagle}        & 12  & 136.9 & 3214 & 163.2 & 9.0  & 121.4 & -    & 421.5 & $1.3\times$ & 0.6618 \\ \midrule
\myname{}                  & 3.5 & 45.1  & 858  & 54.2  & 7.7  & 104.1 & 10.2 & 213.6 & $\mathbf{2.5\times}$ & 0.6602 \\
\myname{} in FP8              & 3.5 & 29.1  & 948  & 39.8  & 6.2  & 84.0  & 10.2 & 163.1 & $\mathbf{3.2\times}$ & 0.6450 \\ \bottomrule
\end{tabular}
\vspace{-0.5cm}
\end{table*}

%% file: impl.tex
\section{Experiments}
\subsection{Dataset}
We evaluate our method using the DriveLM dataset~\cite{DriveLM} with question-answer pairs including reasoning, perception, prediction, and planning capabilities. We randomly sample $150$ scenes exclusively for validation, extracting $100$ QA pairs per scene, resulting in a validation set of $15K$ QA pairs. The remaining data, approximately $210K$ QA pairs, builds up the training set. We ensure no scene overlap between the training and validation sets to guarantee a fair evaluation of generalization performance. We adopt the same evaluation protocol established by DriveLM including GPT scores from GPT-3.5-Turbo~\cite{gpt4v}, language scores (BLEU~\cite{bleu4}, ROUGE-L~\cite{rouge}, and CIDEr~\cite{CIDEr}), and match score. The final score ranging from $0.0$ to $1.0$ is computed as a weighted average of these individual metrics to indicate model's overall performance.

 


\subsection{Training}
We build our pipeline upon InternVL2.5~\cite{InternVL25} 2B and 1B models. For 2B model, it consists of a 300M parameter InternViT encoder and a 1.8B parameter InternLM2.5 decoder; for 1B model, it consists of the same ViT and a 0.5B Qwen2 decoder. We follow the similar process of dataset preparation and training~\cite{mini_internvl}. The training is conducted on 8 NVIDIA A100 GPUs over 15,000 steps, using a per-device batch size of 4. Before training, images from the six surround cameras are individually reshaped to $448$x$996$ pixels and then stitched into a larger composite image(layout: 2 rows of 3 cameras). During training, this composite image is divided into 12 non-overlapping patches of $448$x$448$ each. Based on spatial information inferred from the input query text, only a subset of $12$ patches are selected and fed into the ViT encoder. The draft model learns to predict the candidate tokens using the final hidden state features extracted from our finetuned decoder. 



\subsection{Deployment}
We deploy \textbf{\myname{}} on NVIDIA Drive Thor~\cite{Thor} platform by using a custom build of TensorRT~\cite{tensorrt} to deploy the InternViT encoder, token compression module, InternLM2.5 and Qwen2 decoder, and the Eagle-2 speculative decoding draft model. Post-training quantization (PTQ) for FP8 precision is also tested using ModelOpt~\cite{modelopt} toolkit. 




%% file: evaluation.tex
\subsection{Results}

Figure~\ref{fig:avg-total-latency} compares the average end-to-end latency of our pipeline with baseline $2$B VLMs across different configurations. For the same accuracy target, visual token compression baseline (FastV, K=1) results in a $1.1\times$ reduction; speculative decoding alone offers a $1.3\times$ latency decrease. Our full pipeline achieves a $2.4\times$ end-to-end latency reduction in FP16, which further increases to $3.2\times$ with FP8 quantization, although this comes with slight task accuracy degradation. These figures are based on the average number of generated and input tokens over the validation sets. Table~\ref{table:overall-performance} provides a detailed breakdown of how each component contributes to the overall latency reduction, highlighting the effectiveness of combining patch selection, token compression, and speculative decoding. We also tested the 1B models and achieved a $2.6\times$ overall latency reduction at the combined accuracy of $0.643$.


Regarding memory consumption, in FP16, the ViT engines for both 1B and 2B models require $588.80$ MiB, with a maximum activation memory of $866$ MiB. The $0.5$B Qwen2 engine requires $1.21$ GiB, and its maximum activation is $162$ MiB, while the $1.8$B InternLM2.5 engine requires $3.62$ GiB, with a maximum activation of $288$ MiB. KV-Cache would take $252$ and $864$ MiB for the 1B and 2B models, respectively. The FP8 deployment could potentially reduce the memory consumption by half.


\mypara{Patch Selection Module:} Our Patch Selection Module significantly reduces ViT processing load. On average, processing only $3.5$ strategically selected image patches (down from the default $12$) maintains task performance by filtering out unnecessary views. Figure~\ref{fig:vision-encoder-latency} demonstrates the ViT latency across varying numbers of input patches, confirming a near-linear relationship. This selection results in approximately a $3.0\times$ latency reduction for the vision encoder.

\mypara{Visual Token Compression:} For the LLM prefill stage, which already benefits from patch selection, we apply token compression ratios ($r$) of $0.8$ for FP16 and $0.9$ for FP8 deployments to further minimize prefill latency. In contrast, the baseline FastV requires a more aggressive compression ratio (e.g., $r=0.3$) to achieve similar token counts, which degrades accuracy due to loss of important visual information. As shown in Figures~\ref{fig:first-token-latency} and \ref{fig:extend-one-latency}, while token compression substantially reduces prefill latency, it does not reduce the generation latency much. For example, a $3\times$ reduction of the input tokens only reduces the extend-one-token latency by $1.05$ ms, corresponding to a mere $1.1\times$ speed-up. 

\mypara{Speculative Decoding:} The validation set generates an average of $13.6$ tokens without speculative decoding, this would necessitate a same number of autoregressive iterations, leading to high generation latency. Speculative decoding reduces the number of iterations by nearly $4\times$ on average, accepting $3.8$ tokens per iteration. This significantly lowers generation latency from $16.9 ms$ to $9.0 ms$ on average while maintaining output quality. Furthermore, the average extend-one latency even reduces to $7.7 ms$ and $6.2 ms$ in FP16 and FP8 respectively, since the layers in speculative decoder benefit from the smaller hidden feature sizes from token compression. The overall generation latency is reduced by $2.2\times$ for FP16 and $2.7\times$ for FP8.


\mypara{Quantization:}
Figure~\ref{fig:latency-breakdown} shows the latency speed-ups achieved by FP8 quantization across different VLM execution stages. For the ViT model and the LLM generation stage, FP8 quantization consistently yields speed-ups of approximately $1.5\times$ and $1.6\times$, respectively. The LLM prefill stage also benefits, with speed-ups ranging from $1.3\times$ for smaller numbers of input tokens to $1.7\times$ for larger inputs, averaging a $1.6\times$ speed-up. Note that accuracy loss due to quantization may be mitigated via techniques such as quantization-aware training, and this will be explored in our future work. 

\section{Conclusion}
This work presented a highly efficient Vision-Language Model (VLM) pipeline tailored for latency-critical applications on embedded systems. By integrating three complementary techniques including patch selection, token compression, and speculative decoding, our approach systematically addresses latency bottlenecks at various stages of the inference process. This pipeline was tested on embedded platform with limited resource and significant latency reduction was observed under different settings. These findings advocate the practical deployment of such models on embedded devices, paving the way for their broader adoption in real world.